\documentclass[letterpaper]{article} % DO NOT CHANGE THIS
\usepackage{aaai2026}  % DO NOT CHANGE THIS
\usepackage{times}  % DO NOT CHANGE THIS
\usepackage{helvet}  % DO NOT CHANGE THIS
\usepackage{courier}  % DO NOT CHANGE THIS
\usepackage[hyphens]{url}  % DO NOT CHANGE THIS
\usepackage{graphicx} % DO NOT CHANGE THIS
\usepackage{natbib}  % DO NOT CHANGE THIS AND DO NOT ADD ANY OPTIONS TO IT
\usepackage{caption} % DO NOT CHANGE THIS AND DO NOT ADD ANY OPTIONS TO IT
\usepackage{algorithm}
\usepackage{algorithmic}
\usepackage{amsmath, amsthm, amssymb, amsfonts}  %% Math packages
\usepackage{tabularray}
\usepackage{multirow}
\usepackage{graphicx}
\usepackage{newfloat}
\usepackage{listings}
\DeclareCaptionStyle{ruled}{labelfont=normalfont,labelsep=colon,strut=off} % DO NOT CHANGE THIS
\floatstyle{ruled}
\newfloat{listing}{tb}{lst}{}
\floatname{listing}{Listing}
\title{Out-of-Context Misinformation Detection via Variational Domain-Invariant Learning with Test-Time Training}
\author{
    Xi Yang\textsuperscript{\rm 1},
    Han Zhang\textsuperscript{\rm 1},
    Zhijian Lin\textsuperscript{\rm 1},
    Yibiao Hu\textsuperscript{\rm 1},
    Hong Han \textsuperscript{\rm 1}\thanks{Corresponding author}\\
}
\affiliations{
    \textsuperscript{\rm 1} School of Electronic Engineering, Xidian University, Shaanxi, China\\

     yangxi@stu.xidian.edu.cn, 22021110280@stu.xidian.edu.cn, 23021211252@stu.xidian.edu.cn, yibiaohu@stu.xidian.edu.cn, hanh@mail.xidian.edu.cn 

}

\usepackage{bibentry}
\begin{document}

\maketitle

\begin{abstract}
Out-of-context misinformation (OOC) is a low-cost form of misinformation in news reports, which refers to place authentic images into out-of-context or fabricated image-text pairings.
This problem has attracted significant attention from researchers in recent years. 
Current methods focus on assessing image-text consistency or generating explanations.
However, these approaches assume that the training and test data are drawn from the same distribution. 
When encountering novel news domains, models tend to perform poorly due to the lack of prior knowledge.
To address this challenge, we propose \textbf{VDT} to enhance the domain adaptation capability for OOC misinformation detection by learning domain-invariant features and test-time training mechanisms.
Domain-Invariant Variational Align module is employed to jointly encodes source and target domain data to learn a separable distributional space domain-invariant features. 
For preserving semantic integrity, we utilize domain consistency constraint module to reconstruct the source and target domain latent distribution.
During testing phase, we adopt the test-time training strategy and confidence-variance filtering module to dynamically updating the VAE encoder and classifier, facilitating the model's adaptation to the target domain distribution.
Extensive experiments conducted on the benchmark dataset NewsCLIPpings demonstrate that our method outperforms state-of-the-art baselines under most domain adaptation settings.
\end{abstract}

\begin{links}
    \link{Code}{https://github.com/yanggxii/VDT}
\end{links}

\section{Introduction}

The news reports on online platform is numerous and various.
Particularly, the rapid advancement of generative AI technologies has led to a rapid rise in synthetic data ~\citep{pan2023attacking, chen2023can}, which presents a significant challenge to the authenticity and credibility of news reports.
In contrast to Deepfake model ~\cite{dolhansky2020deepfake} which generates some non-existent images, a low-cost and low-tech type of misinformation is reusing of images ~\citep{jaiswal2017multimedia, fazio2020out, qi2024sniffer}.
This misinformation places the real picture in a out-of-context or a false image-text pair environment, often referred to as out-of-context misinformation(OOC)~\citep{hu2024bad}.
Due to its low technical requirements, misinformation is easy to manipulate, resulting in widely spreading and hard detecting.

\begin{figure}[t!]
    \centering
    \includegraphics[scale=0.48]{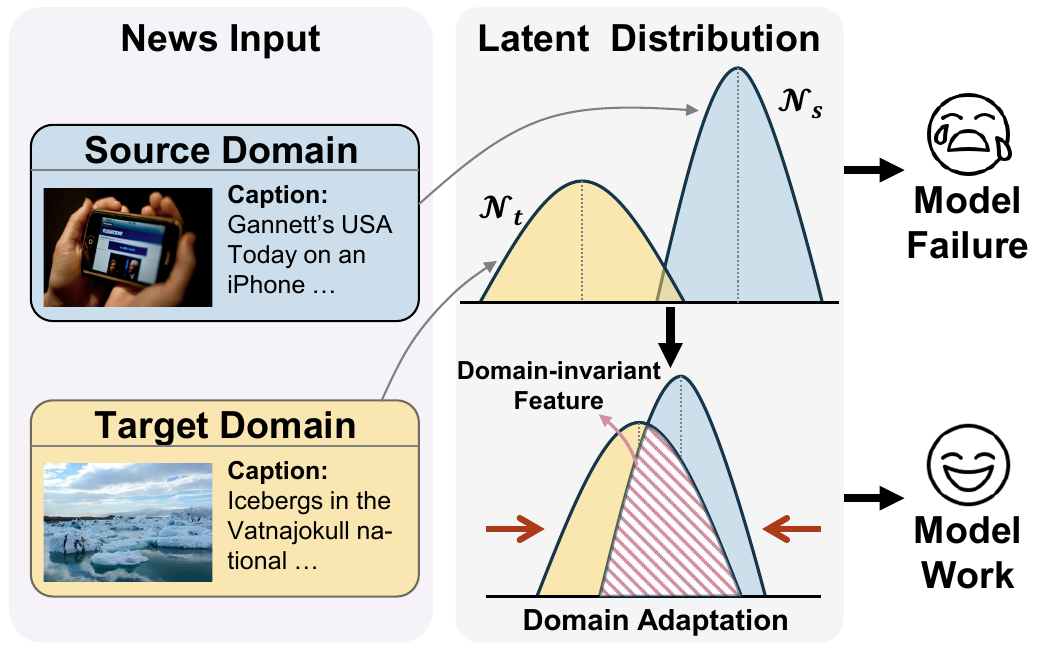}
    \caption{The illustration of domain adaptation challenge in out-of-context misinformation detection. Model failure on target domain due to the significant distribution gap between the source and target domains. By domain adaptation to learn domain-invariant features, the model adapted to target domain better.}
    \label{fig:introduction}
\end{figure}

Faced with this challenge, current research can be divided into two directions roughly.
Some studies tend to calculate the similarity of pictures and text in the representation space ~\citep{jaiswal2017multimedia, luo2021newsclippings, huang2022text} or use external evidence to provide support ~\citep{muller2020multimodal, papadopoulos2023red, papadopoulos2025similarity}.
While other works places greater emphasis on the interpretative side of detection ~\citep{ qi2024sniffer, zhang2023interpretable}.
These approaches leverage Multimodal large language models (MLLMs) to generate plausible and well-grounded explanations that justify the detection outcomes.
Although some progress has been made in these methods, they all assume that the training and testing data are drawn from the same distribution.
In general, news reports encompass a diverse array of topics and domains, with significant variations in thematic style and content across different subjects.
Due to lack of prior knowledge, poor performance of detection models will occur when new news topics emerge.
It is not feasible to retrain the detection model every time a new domain arises, and it is equally difficult to produce datasets for each new topic. 
Figure~\ref{fig:introduction} illustrates the domain adaptation challenge in ooc misinformation detection.

Domain adaptation is an important research direction in transfer learning~\citep{lu2025dammfnd}, aiming to address the fundamental challenge of distribution discrepancy between source and target domain.
Specifically, it enhances model performance on the target domain by utilizing the rich labeled data available in the source domain.
Many studies ~\citep{ganin2016domain, zhang2019bridging, wang2022exploring, tzeng2014deep} have demonstrated that learning domain-invariant features can effectively mitigate the distribution shift between source and target domains.
These features effectively capture the essential information of data in both source and target domain, while remaining robustness to domain-specific attributes and label distributions.
However, in the out-of-context news detection task, there are still some challenges.
First, OOC datasets are limited in scope, covering only a few topics and institutions.
The substantial difference in data characteristics across different domains poses significant challenges for models to learn generalizable domain-invariant features.
Second, the inherent lack of contextual information in manipulated data, which often leads models to over-rely on certain local multimodal features~\citep{geirhos2020shortcut, zhang2021understanding}, thereby exacerbating the difficulty of learning robust domain-invariant representations.

To deal with this challenge, we proposed \textbf{V}ariational \textbf{D}omain-Invariant Learning with \textbf{T}est-Time Training (\textbf{VDT}) for domain adaptive out-of-context news detection.
VDT first uses an Multimodal Large Language Model (MLLM) to directly encode the image and text into a multimodal feature representation.
And then we introduce a domain-invariant variational alignment module that employs a shared VAE to jointly encode the source and target data, learn their latent distributions, and promote feature alignment across domains.
Moreover, we leverage the properties of the mean and variance to enhance the representational capacity of the domain-invariant features.
To reduce the domain gap and ensure that domain-invariant features are captures, we impose constraints on the latent distributions between the source and target domains.
Meanwhile, we employ a domain consistency constraint module to reconstruct the latent distributions to prevent representation collapse.
Finally, we adopt test-time training strategy to enhance the model’s adaptability to the target domain.
During the evaluation phase, a confidence-variance filtering mechanism is applied to screen pseudo-labeled~\citep{lee2013pseudo} that from unseen target domain.
High-quality samples are retained to update the VAE encoder and the classifier.

The main contributions of our work are four-fold:
\begin{itemize}
    \item We propose a variational domain-invariant learning with test-time training framework to address the performance degradation caused by distribution shifts between the source and target domains in out-of-context misinformation detection, thereby improving the model’s domain adaptation capability.
    
    \item To prevent the model from relying on multimodal local features due to inherent lack of contextual information in manipulated data, we introduce the DIVA module to learn latent distributions. By imposing constraints to reduce the domain gaps, ensuring that domain-invariant features are effectively captured.
    
    \item During the test phase, we adopt a test-time training strategy to dynamically update the VAE encoder and classifier. confidence-variance filtering mechanism is employed to select high-quality samples, thereby model can better adapt to the target domain.
    
    \item Extensive experiments on the widely-used benchmark datasets, NewsCLIPpings, demonstrate that our proposed method achieves strong and competitive performance under various domain adaptation settings.
\end{itemize}

\section{Related Work}

\subsection{Domain Adaptation News Detection}
Current research primarily focuses on evaluating image-text consistency~\citep{jaiswal2017multimedia, aneja2021cosmos, luo2021newsclippings, mu2023self, papadopoulos2023red, yuan2023support} and leveraging large language models to enhance interpretability~\citep{papadopoulos2023red, zhang2023ecenet, papadopoulos2025similarity}.
But the domain adaptation challenges to out-of-context misinformation detection is neglected.
Many studies have adopted adversarial learning~\citep{wang2018eann, yuan2021improving, zhang2020multimodal}  to extract domain-invariant features. 

Alternative research directions have explored discrepancy-based approaches~\citep{yue2022contrastive, gu2024learning}, which achieve domain adaptation by minimizing the distribution divergence between source and target domains.

\begin{figure*}[t!]
    \centering
    \includegraphics[scale=0.58]{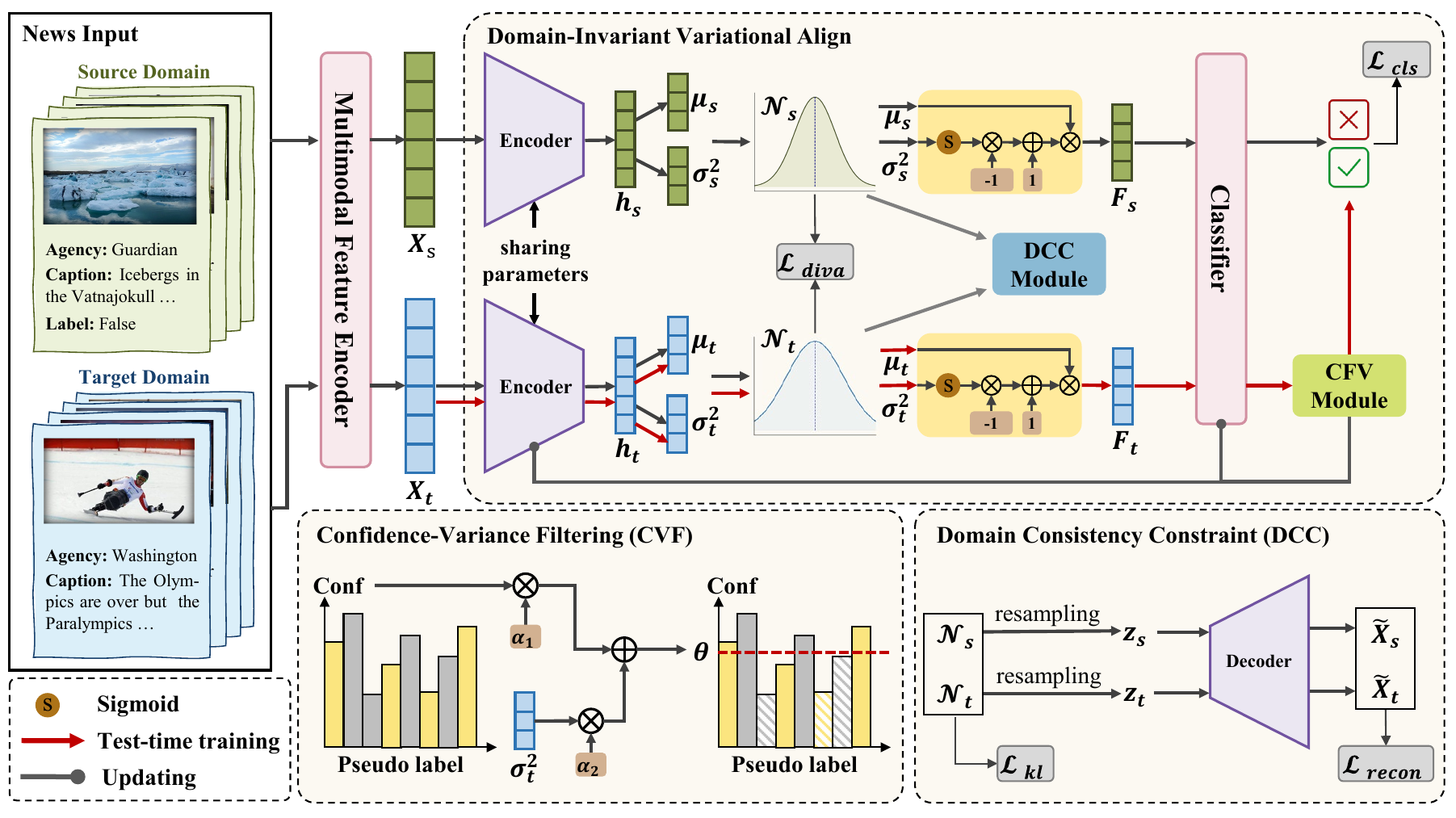}
    \caption{The illustration of proposed VDT model.}
    \label{figures:method}
\end{figure*}

\subsection{Test-Time Training}
Test-Time Training methods~\citep{ wu2024test} update the model at inference time using a Y-shaped architecture, in which the main task and a self-supervised auxiliary task are jointly optimized during training.

Sun et al.~\cite{osowiechi2023tttflow, sun2020test} introduced a seminal technique that later gave rise to a family of methods collectively known as Test-Time Training (TTT), which enables re-training the network on the test set even though no test labels are available.
The original work employed rotation prediction~\citep{komodakis2018unsupervised} as the auxiliary task.
Later approaches~\citep{gandelsman2022test, liu2021ttt++} replaced the original auxiliary task with either Masked Autoencoder reconstruction~\citep{he2022masked} or contrastive learning~\citep{chen2020simple}.

In our work, we adopt VAE reconstruction as the auxiliary task, and update the VAE encoder during the TTT phase.
The decoder is kept frozen to prevent it from being influenced by target domain samples, which may distort the original reconstruction structure.
Moreover, since the target domain data are not seen by the classifier during training, the classifier may exhibit a source-domain bias.
To mitigate this, we also update the classifier parameters during test-time training.

\section{Method}

\subsection{Multimodal Feature Encoder}
Following Gu et al.~\citep{gu2024learning}, we use MLLM to encode the news text and image into a unified semantic representation space, for capturing the semantic differences between original news and out-of-context news.
It enables the model to learn rich cross-modal interactions, facilitating more accurate detection of semantic inconsistencies.
Formally, the multimodal feature encoder is defined as follows:
\begin{equation}
\begin{aligned}
X^s = MLLM((x_{img}, x_{txt})^s)\\
X^t = MLLM((x_{img}, x_{txt})^t)
\end{aligned}
\end{equation}
where $X^S$ and $X^T$ denote the multimodal features representation of the news image-text pair $(x_{img}, x_{txt})$ from the source and target domain.

\subsection{Domain-Invariant Variational Align}
After encoding the image-text pairs in news articles, we introduce a Domain-Invariant Variational Align (DIVA) module to model the latent distributions of source and target domain samples.

In DIVA module, we adopt Variational Autoencoder (VAE)~\citep{kingma2013auto, park2024data}, whose generative capacity facilitates the learning of separable latent features for both the source and target domain distributions.
Specifically, the latent distributions of source and target domains are learned through two VAE encoders that share parameters.
The process is defined as follows:
\begin{equation}
\begin{aligned}
h_s = E_s(X_s; \theta^e) \qquad
h_t = E_t(X_t; \theta^e)
\end{aligned}
\end{equation}
where $\theta^e$ is denotes the learnable parameters of the encoder.

The latent representations of the source $h_S$ and target domains $h_t$ are further passed through distinct linear layers to obtain the mean and variance respectively, as follows:
\begin{equation}
\begin{aligned}
\mu_s = W_{s\mu}h_s + b_{s\mu} \qquad
log\sigma_s^2 = W_{s\sigma}h_s + b_{s\sigma}
\end{aligned}
\end{equation}
where $\mu_s$ is the mean of the latent distribution for the source domain, $log\sigma_s^2$ represent the variance. 
$W_{s\mu}$, $b_{s\mu}$ denote the weights and bias of the mean layer, and $W_{s\sigma}$, $b_{s\sigma}$ represent the weights and bias of the log-variance layer.

Similarly, the mean and variance of the target domain can be obtained in the same manner, represent as $\mu_t$ and $\sigma^2_t$.
In this way, we obtain Gaussian distributions in the latent space for both the source and target domains, denoted as: $\mathcal{N}_s \sim (\mu_s, \sigma^2_s)$ and $\mathcal{N}_t \sim (\mu_t, \sigma^2_t)$.

The mean $\mu$ represents the semantic center of samples in the latent space, while the variance $\sigma^2$ controls the degree of dispersion.
A larger variance indicates that the sample is more spread out, farther from the semantic center, and associated with higher uncertainty.

To construct more robust domain-invariant features, we transform the variance into gating weights and express the final domain-invariant feature as:
\begin{equation}
\begin{aligned}
F_s = \mu_s(1-gate_s) \qquad
gate_s = sigmoid(log\sigma^2_s)
\end{aligned}
\end{equation}

This representation ensures that samples with high uncertainty are attenuated in weight, effectively reducing the influence of features far from the semantic center.
Conversely, samples with low variance retain their weights.

Contrastive learning has been shown to effectively pull semantically similar samples closer while pushing dissimilar ones apart in the latent space~\citep{bhattacharjee2023conda}.
We adopt a contrastive loss to constrain the mean representations of the source and target domains, aiming to align their distributions and reduce the domain gap.
We refer to this objective as the DIVA loss, which is formulated as follows:
\begin{equation}
\begin{aligned}
\mathcal{L}_{diva} = -log \sum_{i \in N} \frac{exp(sim(\mu_{si}, \mu_{ti})/ \tau)}{\sum^{2N}_{j=1,j\ne i}exp(sim(\mu_{sj}, \mu_{tj})/ \tau)}
\end{aligned}
\end{equation}
where $N$ denotes the batch size, and $\mu_{si}$ and $\mu_{ti}$ represent the mean vectors of the $i$-th sample from the source and target domains, respectively. $\tau$ is a temperature parameter, and $sim(\cdot)$ denotes the similarity metric, for which we use cosine similarity.

\subsection{Domain Consistency Constraint Module}
After obtaining the latent distributions of the source and target domains, we employ the reparameterization trick to enable gradient-based optimization.
\begin{equation}
\begin{aligned}
z_s = \mu_s + \sigma_s \cdot \epsilon, \quad \epsilon\in\mathcal{N}(0, I) \\
z_t = \mu_t + \sigma_t \cdot \epsilon, \quad \epsilon\in\mathcal{N}(0, I)
\end{aligned}
\end{equation}
where random noise $\epsilon$ is sampled from a normal distribution $\mathcal{N}(0, I)$.

To prevent the encoder from losing critical information while aligning the source and target domains—i.e., to avoid semantic collapse caused by over-alignment. We pass the sampled latent codes $z_s$ and $z_t$ through a shared decoder to reconstruct the original inputs. Formally, this can be expressed as:
\begin{equation}
\begin{aligned}
\hat{X_s} = D_s(z_s; \theta^d) \qquad
\hat{X_t} = D_t(z_t; \theta^d)
\end{aligned}
\end{equation}
where $\theta^d$ is denotes the learnable parameters of the decoder.

We formulate the reconstruction loss ensuring the distributions effectively capture critical semantic features and maintain semantic fidelity,  as follows:
\begin{equation}
\begin{aligned}
\mathcal{L}_{recon} = ||X_s-\hat{X}_s||^2 + ||X_t-\hat{X}_t||^2
\end{aligned}
\end{equation}

As training epochs number increases, the encoder progressively compresses the data into an extremely narrow region of the latent space, diminished expressiveness of the latent variables.
To mitigate such representational degradation, we introduce a Kullback–Leibler (KL) divergence~\citep{hershey2007approximating} loss as a regularization term to constrain the latent distributions of both the source and target domains toward a standard normal distribution $\mathcal{N}(0, I)$. 
This way encourages the target domain representations to be implicitly aligned with those of the source domain, thereby promoting domain alignment and improving the coherence and sampleability of the learned latent space.
The KL divergence loss is defined as:
\begin{equation}
\begin{aligned}
\mathcal{L}_{kl} &= \frac{1}{2} \sum_{m \in \{ s,t\}} KL(\mathcal{N}(\mu_m, \sigma^2_m)\ ||\ \mathcal{N}(0,I)) \\ 
 & = \frac{1}{2} \sum_{m \in \{ s,t\}} \frac{1}{2}(\mu_m^2 + \sigma_m^2 - log\sigma_m^2 -1)
\end{aligned}
\end{equation}

Therefore, the overall loss of domain consistency constraint module can be expressed as:
\begin{equation}
\begin{aligned}
\mathcal{L}_{dcc} = \mathcal{L}_{recon} + \beta \mathcal{L}_{kl}
\end{aligned}
\end{equation}
where the $\beta$ parameter is used to weight the Kullback-Leibler (KL) divergence term in the Evidence Lower Bound (ELBO) objective~\citep{burgess2018understanding}. It serves as a critical hyperparameter that balances reconstruction fidelity and latent disentanglement in the VAE framework.

\subsection{Training}
The final domain-invariant feature $F$ is used for classification to obtain the predicted labels of out-of-context news instances, formulated as:
\begin{equation}
\begin{aligned}
\hat{y} = \arg\max_{c}(f) \qquad f = CLS(F)
\end{aligned}
\end{equation}

The loss function for the classifier is the cross-entropy loss, where $\hat{y}$ is the predicted label and $y$ is the ground truth label:
\begin{equation}
\begin{aligned}
\mathcal{L}_{cls} = -E_{ y\sim Y}[ylog(\hat{y}) + (1-y)log(1-\hat{y})]
\end{aligned}
\end{equation}

We combine classification loss $\mathcal{L}_{cls}$, the domain-invariant variational align loss $\mathcal{L}_{diva}$, and the domain consistency constraint loss $\mathcal{L}_{dcc}$ as optimization objective.
The overall loss is:
\begin{equation}
\begin{aligned}
\mathcal{L} = \lambda_1\mathcal{L}_{cls} + \lambda_2\mathcal{L}_{diva} + \lambda_3\mathcal{L}_{dcc}
\end{aligned}
\end{equation}

\subsection{Test-Time Training}
To further adapt the model to the target domain, we introduce a test-time training strategy that leverages the pre-trained network to adjust its parameters using the unlabeled target domain.

First, we freeze all model parameters except for the VAE encoder and the classifier.
Then, the DIVA module is utilized to obtain the mean and variance of the target domain test samples, from which the domain-invariant feature $F$ is derived, formulated as:
\begin{equation}
\begin{aligned}
F_t = \mu_t(1-sigmoid(log\sigma^2_t))
\end{aligned}
\end{equation}

The pseudo-labels and confidence scores are obtained through the classifier as follows:
\begin{equation}
\begin{aligned}
\widetilde{y} = \arg\max_{c}(f_t) \qquad  f_t = CLS(F_t)
\end{aligned}
\end{equation}

\begin{equation}
\begin{aligned}
conf = \max_{c}(f)
\end{aligned}
\end{equation}

The confidence score $\text{conf}$ corresponds to the maximum value among the two-dimensional predicted probabilities, indicating the model's degree of certainty in its current prediction.
\subsubsection{Confidence-Variance Filtering Module} 

To further improve the quality of pseudo-labels, we design a Confidence-Variance Filtering (CVF) Mechanism.

At this stage, we have access to the variance features from the representation layer and the confidence scores for the unlabeled target domain samples.
The variance reflects the representational reliability of a sample—whether its features are stable and can effectively characterize its semantics.
The confidence score indicates the predictive reliability,  that is, the degree of certainty of the model's classification decision.

Only when the samples simultaneously meet the conditions of "high confidence + low variance" are selected as high-quality pseudo-label samples. 

This mechanism can be specifically expressed as:
\begin{equation}
\begin{aligned}
score = \alpha_1 \cdot conf + \alpha_2 \cdot \sigma^2_t
\end{aligned}
\end{equation}
And set a threshold $\theta$. 
We only retain the data with scores higher than the threshold $\theta$ to update the VAE encoder and classifier.

\subsubsection{Updating}
The filtered data is reconstructed to obtain $\widetilde{X}_t$.
During test-time training, the loss function consist of the cross-entropy loss, reconstruction loss, and KL divergence loss, which is expressed as:
\begin{equation}
\begin{aligned}
\mathcal{L}_{TTT} = \mathcal{L}_{cls} + \mathcal{L}_{dcc}
\end{aligned}
\end{equation}

\section{Experiments}
\subsection{Dataset and Evaluation Metrics}
\subsubsection{Dataset}

\begin{table}[]
\begin{tabular}{ccccc}
\hline
\multicolumn{1}{c}{\multirow{2}{*}{Agency}} & \multicolumn{2}{c}{train}                               & \multicolumn{2}{c}{test} \\ \cline{2-5} 
\multicolumn{1}{c}{}                        & \multicolumn{1}{c}{class0} & \multicolumn{1}{c}{class1} & class0      & class1     \\ \hline
Guardian                                    & 283782                     & 283230                     & 29631       & 29532      \\
BBC                                         & 108060                     & 97908                      & 11561       & 10513      \\
Washington Post                             & 89914                      & 91613                      & 9217        & 9471       \\
USA Today                                   & 109694                     & 118699                     & 11675       & 12568      \\ \hline
\end{tabular}
\caption{The statistics of NewsCLIPpings dataset. The four quantities in each agency are respectively train set-class0, train Set-class1, test set-class0, and test set-class1.}
\label{tab:dataset statistic}
\end{table}

We conduct experiments on NewsCLIPpings dataset~\citep{luo2021newsclippings}, which is collected from VisualNews dataset~\citep{liu2021visual}, a benchmark news image caption dataset. 
The sources of the news include four institutions: Guardian (G), BBC (B), Washington Post (W) and USA Today (U).
The labels are balanced, and the distribution of samples across different news agencies is detailed in Table~\ref{tab:dataset statistic}.
In this dataset, we treat different news agencies as distinct domains to address the domain adaptation problem for OOC misinformation detection.

\subsubsection{Evaluation Metrics}
For evaluation metrics, we report accuracy and macro $F1$ score.
Alongside the main metrics, we further include $F1_{real}$ and $F1_{fake}$~\citep{hu2024bad} to better understand class-wise performance.

\subsection{Implementation details}
The proposed VDT model is implemented in PyTorch 1.13.1 with CUDA 11.6, and all experiments are run on a single NVIDIA RTX 3080 Ti GPU. 
And We use BLIP-2’s multimodal feature extractor~\citep{li2023blip} to embed the image and text, and obtain the embedding $x$ of size 768.
We employ the Adam optimizer for training with a batch size of 256 for 20 epochs.
The early stopping epoch is set to 5.
In training stage, the learning rate is set to $\mathrm{1e-4}$ and in test-time training phase, the learning rate is set to $\mathrm{2e-5}$.
The dim of $\mu$ and $\sigma^2$ is set to 128, and the dim of classifier layer is set to 500.

In the loss function, the weight of kl divergence loss $\beta$ is set to 1.5. 
$\lambda_1$, $\lambda_2$, and $\lambda_3$ is set to 2, 0.5 and 1.
The $\alpha_1$ is set to 2 and the $\alpha_2$ is set to -1.
The threshold $\theta$ is set to 0.9.

For NewsCLIPpings dataset, inspired by Gu et al.~\citep{gu2024learning}, we respectively used two news organizations from the same country as the target domain and the remaining two news organizations as the source domain.
During the training phase, we have access to labeled source domain data and unlabeled target domain data.
In the testing phase, we evaluate the model's performance on the labeled target domain data.

\begin{table*}[t!]
\setlength\tabcolsep{2pt} 
\begin{tabular}{cccccccccccccccccccc}
\hline
\multicolumn{1}{c}{\multirow{2}{*}{Model}} & \multicolumn{4}{c}{U, W $\rightarrow$ B}        & \multicolumn{1}{c}{} & \multicolumn{4}{c}{U, W $\rightarrow$ G}   & \multicolumn{1}{c}{} & \multicolumn{4}{c}{B, G $\rightarrow$ U}                        & \multicolumn{1}{c}{} & \multicolumn{4}{c}{B, G $\rightarrow$ W}   \\ \cline{2-5} \cline{7-10} \cline{12-15} \cline{17-20} 
\multicolumn{1}{c}{}                       & $F1$             & Acc            & $F_{real}$ & $F_{fake}$ &                      & $F1$             & Acc            & $F_{real}$ & $F_{fake}$ &                      & $F1$             & Acc            & $F_{real}$ & $F_{fake}$ &                      & $F1$             & Acc            & $F_{real}$ & $F_{fake}$ \\ \hline
EANN                                       & 68.72          & 69.82          & 65.18   & 72.26   &                      & 75.93          & 76.05          & 74.30   & 77.56   &                      & 79.80          & 79.70          & 78.06   & 81.53   &                      & 78.33          & 78.33          & 77.07   & 79.58   \\
MDA-WS                                     & 70.02          & 69.99          & 72.20   & 67.85   &                      & 74.95          & 75.23          & 72.39   & 77.52   &                      & 79.51          & 79.46          & 77.64   & 81.39   &                      & 78.14          & 78.14          & 76.92   & 79.36   \\
CANMD                                      & 68.61          & 68.53          & 66.70   & 71.37   &                      & 74.78          & 75.06          & 74.01   & 77.79   &                      & 80.30          & 80.29          & 78.67   & 81.85   &                      & 78.60          & 78.50          & 77.40   & 78.62   \\
REAL-FND                                   & 69.61          & 69.84          & 66.81   & 72.42   &                      & 75.39          & 75.59          & 73.14   & 77.63   &                      & 80.33          & 80.45          & 78.78   & 81.87   &                      & \underline{78.75}    & \underline{78.77}    & 78.10   & 79.41   \\
CADA                                       & 68.73          & 69.85          & 65.12   & 72.34   &                      & \underline{75.65}    & \underline{75.83}    & 73.66   & 77.64   &                      & 80.31          & 80.26          & 78.83   & 81.80   &                      & 78.27          & 78.27          & 77.24   & 79.31   \\
ConDA-TTT                                  & \underline{71.86}    & \underline{71.83}    & 73.78   & 69.94   &                      & 75.40          & 75.66          & 78.02   & 72.79   &                      & \underline{81.03}    & \underline{81.34}    & 81.87   & 80.18   &                      & 78.61          & 78.74          & 79.62   & 77.61   \\ \hline
VDT                                        & \textbf{72.26} & \textbf{72.30} & 74.08   & 70.26   &                      & \textbf{76.59} & \textbf{76.63} & 77.53   & 75.65   &                      & \textbf{81.96} & \textbf{81.95} & 81.38   & 82.49   &                      & \textbf{79.02} & \textbf{79.02} & 78.61   & 79.42   \\ \hline
\end{tabular}
\caption{Performance (\%) comparison between VDT and baselines on NewsCLIPpings. B, G, U and W stand for BBC, Guardian, USA Today and Washington Post respectively. In X $\rightarrow$ Y, X denotes the source domain, Y denotes the target domain. The best performance is in bold text. The second best performance is underlined.}
\label{tab: maim results}
\end{table*}

\subsection{Compare to the State-Of-The-Art}
We conduct a fair and reasonable comparison between VDT and six representative, republic state-of-the-art(SOTA) methods, including: EANN~\citep{wang2018eann}, MDA-WS~\citep{li2021multi}, CANMD~\citep{yue2022contrastive}, REAL-FND~\citep{mosallanezhad2022domain}, CADA~\citep{mosallanezhad2022domain}, and ConDA-TTT~\citep{gu2024learning}.
On the NewsCLIPpings dataset, our proposed VDT outperforms baseline models under the vast majority of domain adaptation settings.
Specifically, when targeting domain G (Guardia), VDT achieves improvements of 1.19\% in $F1$ score and 0.97\% in accuracy compared to ConDA-TTT.
Similarly, when domain U (USA Today) is used as the target, our method surpasses the strongest baseline by 0.93\% in $F1$ score and 0.61\% in accuracy.

Experimental results demonstrate that our proposed model effectively aligns the distributions of the source and target domains, which reduces the domain gap and enables the extraction of discriminative domain-invariant features, thereby enhancing the model's generalization performance on the target domain.

\subsection{Different Domain Adaptation Setting}

\begin{table}[]
\centering
\begin{tabular}{cccccc}
\hline
\multicolumn{1}{c}{\multirow{2}{*}{Mode}} & \multicolumn{2}{c}{ConDA-TTT} & \multicolumn{1}{c}{} & \multicolumn{2}{c}{VDT} \\ \cline{2-3} \cline{5-6} 
\multicolumn{1}{c}{}                      & F1            & Acc           &                      & F1         & Acc        \\ \hline
B$\rightarrow$U          & 77.16         & 77.63         &                      & 75.96      & 76.04      \\
U$\rightarrow$B          & 71.23         & 71.19         &                      & 71.55      & 71.60      \\
B$\rightarrow$W          & 76.68         & 76.92         &                      & 76.84      & 76.86      \\
W$\rightarrow$B           & 72.01         & 71.99         &                      & 71.75      & 71.75      \\
B$\rightarrow$G             & 73.59         & 74.16         &                      & 74.55      & 74.56      \\
G$\rightarrow$B             & 73.27         & 73.23         &                      & 72.63      & 72.62      \\
G$\rightarrow$U              & 81.34         & 81.59         &                      & 81.65      & 81.66      \\
U$\rightarrow$G              & 75.07         & 75.31         &                      & 73.95      & 74.40      \\
G$\rightarrow$W              & 79.00         & 79.11         &                      & 79.12      & 79.12      \\
W$\rightarrow$G            & 74.73         & 74.98         &                      & 75.79      & 75.96      \\
U$\rightarrow$W            & 79.30         & 79.33         &                      & 79.27      & 79.26      \\
W$\rightarrow$U            & 82.27         & 82.49         &                      & 83.28      & 83.27      \\ \hline
GUW$\rightarrow$B         & 72.18         & 72.15         &                      & 71.85      & 71.84      \\
BUW$\rightarrow$G          & 75.33         & 75.45         &                      & 77.14      & 77.15      \\
BGW$\rightarrow$U          & 80.42         & 80.74         &                      & 82.78      & 82.77      \\
BGU$\rightarrow$W          & 78.52         & 78.60         &                      & 79.76      & 79.78      \\ \hline
ALL$\rightarrow$B         & 74.01         & 73.96         &                      & 74.72      & 74.86      \\
ALL$\rightarrow$G         & 76.64         & 76.81         &                      & 78.21      & 78.21      \\
ALL$\rightarrow$U          & 82.77         & 83.04         &                      & 84.52      & 84.51      \\
ALL$\rightarrow$W          & 80.18         & 80.31         &                      & 80.73      & 80.74      \\ \hline
\end{tabular}
\caption{Performance (\%) of different domain adaptation settings on NewsCLIPpings. In X $\rightarrow$ Y, X denotes the source domain, Y denotes the target domain. "ALL" include four agencies (B, G, U, and W).}
\label{tab:dfferent DA setting}
\end{table}

\begin{table}[]
\setlength\tabcolsep{3pt} 
\begin{tabular}{ccccccc}
\hline
\multicolumn{2}{c}{}            & VDT   & $w/o$ $\mathcal{L}_{diva}$ & $w/o$ $\mathcal{L}_{dcc}$ & $w/o$ TTT & $w/o$ CVF \\ \hline
\multirow{2}{*}{B} & $F1$  & 72.26 & 68.87       & 62.80      & 69.08   & 68.13   \\
                          & Acc & 72.30 & 69.12       & 63.87      & 69.25   & 67.77   \\ \hline
\multirow{2}{*}{G} & $F1$  & 76.59 & 73.71       & 68.41      & 75.30   & 74.27   \\
                          & Acc & 76.63 & 74.29       & 69.36      & 75.43   & 74.57   \\ \hline
\multirow{2}{*}{U} & $F1$  & 81.96 & 79.82       & 72.06      & 79.57   & 79.84   \\
                          & Acc & 81.95 & 79.90       & 72.27      & 79.72   & 79.85   \\ \hline
\multirow{2}{*}{W} & $F1$  & 79.02 & 78.23       & 70.87      & 77.98   & 77.27   \\
                          & Acc & 79.02 & 78.30       & 71.86      & 77.99   & 77.43   \\ \hline
\end{tabular}
\caption{Ablation study on NewsCLIPpings dataset. $w/o$ indicates the ablation of the corresponding component. B, G, U, and W denote the target domains, and the experimental settings are consistent with those in Table~\ref{tab: maim results}.}
\label{tab:Ablation}
\end{table}

\begin{figure}[t!]
    \centering
    \includegraphics[scale=0.38]{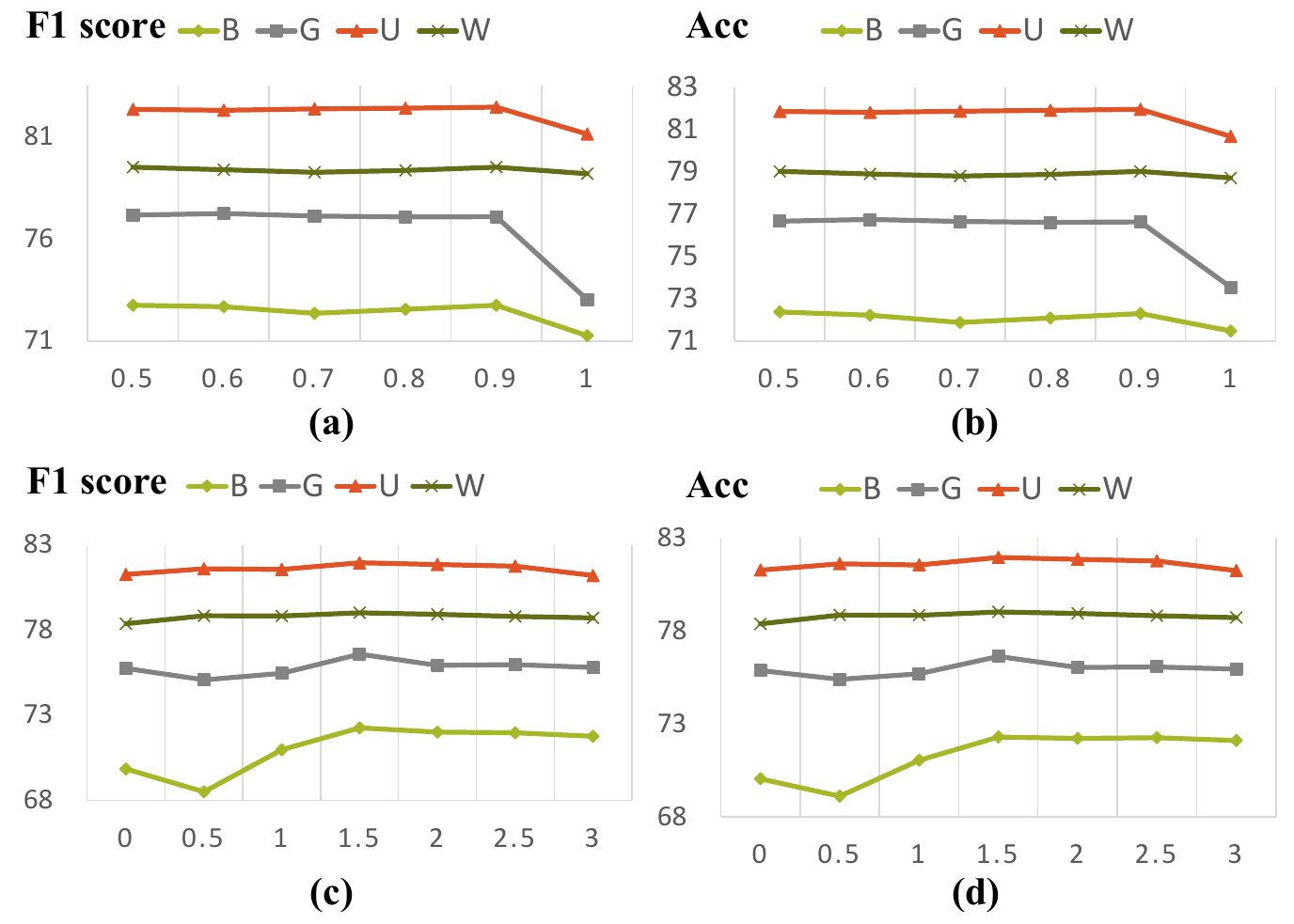}
    \caption{VDT’s performances with different $\beta$ (sub-figure (a) and (b)) and threshold $\theta$ (sub-figure (c) and (d)) values. The legend illustrates target domains.}
    \label{fig:sensitive}
\end{figure}

To complement the main experiments and cover domain adaptation settings not previously included, we conduct additional experiments to further validate the effectiveness of the proposed method in domain adaptation tasks. 
The results are shown in Table~\ref{tab:dfferent DA setting}.

Table~\ref{tab:dfferent DA setting} includes three scenes: using a single domain, three domains, and all domains as the source domain.
The experimental results show that in the vast majority of domain adaptation settings, our proposed method consistently outperforms the state-of-the-art model ConDA-TTT.
In particular, for the BGW→U setting, our method outperforms ConDA-TTT by 2.36\% in $F1$ score and 2.03\% in accuracy.
Significant performance differences between B$\rightarrow$U \& U$\rightarrow$B and G$\rightarrow$U \& U$\rightarrow$G indicate a strong asymmetry between these domains.
The adaptive direction has a notable impact on performance~\citep{farahani2021brief, you2019universal}, likely due to the higher diversity or quality of source domains (e.g., G), which makes them more effective as source domains. Conversely, target domains like U may have simpler data distributions.

Furthermore, we observe that multi-source adaptation generally performs better than single-source adaptation. For example, ALL$\rightarrow$U achieves better performance than G$\rightarrow$U and W$\rightarrow$U.
Our proposed method performs particularly well in multi-source settings, demonstrating its superior effectiveness and generalization capability.

\begin{figure*}[t!]
    \centering
    \includegraphics[scale=0.47]{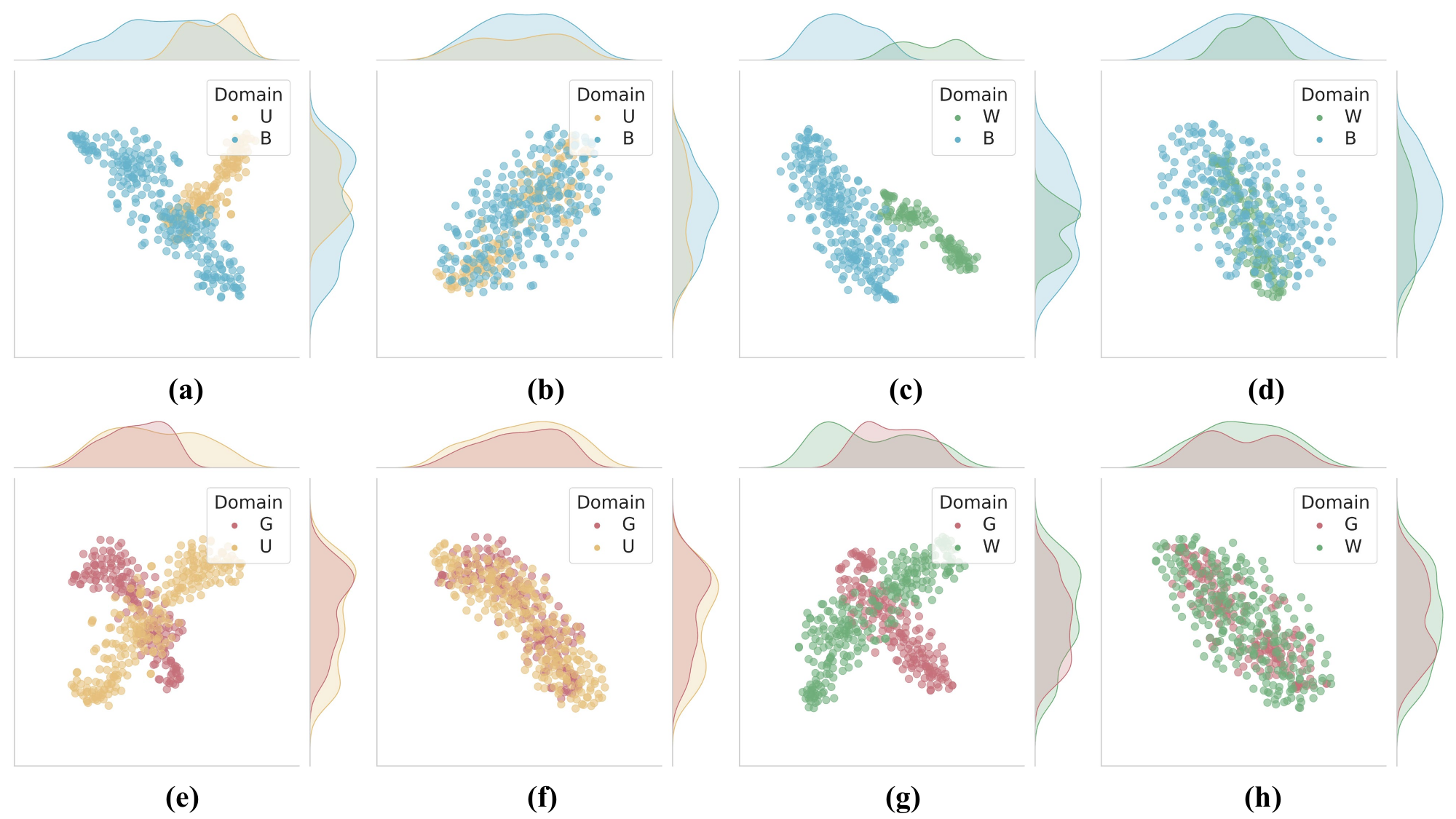}
    \caption{t-SNE visualization of the multimodal feature $X$ and the learned domain-invariant feature $F$.}
    \label{fig:visualization}
\end{figure*}

\subsection{Ablation Study}

We conduct an ablation study on different components of VDT to verify the effectiveness of each proposed module, and the results are shown in Table~\ref{tab:Ablation}.
Specifically, we ablate $\mathcal{L}{diva}$, $\mathcal{L}{dcc}$, the CVF mechanism, and TTT.

From the experimental results, we observe that removing any of these components leads to varying degrees of performance degradation.
In particular, ablating $\mathcal{L}{dcc}$ causes the most significant drop in performance.
This is mainly because $\mathcal{L}{dcc}$ is responsible for the VAE reconstruction task.
Without it, the latent space lacks proper constraints, leading to encoder degeneration and loss of representational capacity.
This fundamentally disrupts the learning of domain-invariant features.
The performance drop caused by removing the CVF mechanism indicates that low-quality samples have a substantial negative impact on model performance.

\subsection{Sensitivity analysis}
The DCC module and Test-Time Training (TTT) strategy are two key components of our proposed method.
In this section, we investigate how the hyperparameter $\beta$ in the $\mathcal{L}_{\text{dcc}}$ loss and the threshold $\theta$ in TTT influence the performance of VDT.

Figure 3 illustrates how VDT’s performance changes as $\beta$ increases from 0 to 3 with a step size of 0.5.
The results show that the model achieves the best performance across different domain adaptation settings when $\beta$ = 1.5.
When $\beta < 1$, the model prioritizes reconstruction accuracy over disentanglement, making it difficult to learn a separable representation space.
As $\beta$ approaches 3, performance degrades because the model tends to ignore semantic reconstruction, which may result in representation collapse.

In addition, we evaluate the impact of the confidence threshold $\theta$ used in the CVF mechanism.
We test values of $\theta$ from 0.5 to 1, and the model performance under different target domain settings is illustrated in Figure~\ref{fig:sensitive}.
The results show that the model achieves the best performance when $\theta$ = 0.9, indicating that samples with higher confidence scores tend to provide more reliable pseudo-labels.
However, when the threshold exceeds 0.9, the performance drops sharply, likely due to the significant reduction in the number of selected samples.

\subsection{Visualization of domain-invariant feature}

To further investigate whether the DIVA module in VDT effectively aligns the source and target domains and facilitates the learning of domain-invariant features.
We employ t-SNE~\citep{van2008visualizing, dimitriadis2018t} to project both the original multimodal features $X$ and the learned domain-invariant features $F$ into a 2D Euclidean space. 
The resulting visualizations are presented in Figure~\ref{fig:visualization}.

Figure~\ref{fig:visualization} consists of eight subfigures grouped into four pairs: (a)-(b), (c)-(d), (e)-(f), and (g)-(h), where each pair compares the original multimodal features $X$ and the domain-invariant features $F$ learned through the DIVA module.
A comparison within each pair reveals that the originally dispersed source and target domain features become highly overlapped in the projected space after passing through the DIVA module.
This observation indicates that the distribution gap between the two domains is significantly reduced, demonstrating the effectiveness of the DIVA module in aligning cross-domain distributions and extracting domain-invariant representations.

\section{Conclusion}
In conclusion, we proposed VDT, a domain adaptive out-of-context news detection model that can aligns two domains with different distributions to adapt to the unseen target domain.
The proposed approach introduces the DIVA module to jointly encode the latent distributions of source and target domains, along with a latent space alignment loss that effectively facilitates the learning of domain-invariant features.
To prevent semantic collapse, a domain consistent constraint is applied to reconstruct the latent distributions.
During inference, we develop CVF mechanism combined with a test-time training strategy to dynamically adapt to the target domain distribution.
Extensive experiments demonstrate that our method consistently outperforms existing state-of-the-art approaches across various cross-domain settings, showing strong generalization and robustness.

\bibliography{aaai2026}

@inproceedings{pan2023attacking,
  title={Attacking Open-domain Question Answering by Injecting Misinformation},
  author={Pan, Liangming and Chen, Wenhu and Kan, Min-Yen and Wang, William Yang},
  booktitle={Proceedings of the 13th International Joint Conference on Natural Language Processing and the 3rd Conference of the Asia-Pacific Chapter of the Association for Computational Linguistics (Volume 1: Long Papers)},
  pages={525--539},
  year={2023}
}

@inproceedings{chen2023can,
  title={Can LLM-Generated Misinformation Be Detected?},
  author={Chen, Canyu and Shu, Kai},
  booktitle={NeurIPS 2023 Workshop on Regulatable ML},
  year={2023}
}

@article{dolhansky2020deepfake,
  title={The deepfake detection challenge (dfdc) dataset},
  author={Dolhansky, Brian and Bitton, Joanna and Pflaum, Ben and Lu, Jikuo and Howes, Russ and Wang, Menglin and Ferrer, Cristian Canton},
  journal={arXiv preprint arXiv:2006.07397},
  year={2020}
}

@article{yuan2021improving,
  title={Improving fake news detection with domain-adversarial and graph-attention neural network},
  author={Yuan, Hua and Zheng, Jie and Ye, Qiongwei and Qian, Yu and Zhang, Yan},
  journal={Decision Support Systems},
  volume={151},
  pages={113633},
  year={2021},
  publisher={Elsevier}
}

@article{geirhos2020shortcut,
  title={Shortcut learning in deep neural networks},
  author={Geirhos, Robert and Jacobsen, J{\"o}rn-Henrik and Michaelis, Claudio and Zemel, Richard and Brendel, Wieland and Bethge, Matthias and Wichmann, Felix A},
  journal={Nature Machine Intelligence},
  volume={2},
  number={11},
  pages={665--673},
  year={2020},
  publisher={Nature Publishing Group UK London}
}

@inproceedings{zhang2019bridging,
  title={Bridging theory and algorithm for domain adaptation},
  author={Zhang, Yuchen and Liu, Tianle and Long, Mingsheng and Jordan, Michael},
  booktitle={International conference on machine learning},
  pages={7404--7413},
  year={2019},
  organization={PMLR}
}

@inproceedings{wang2022exploring,
  title={Exploring domain-invariant parameters for source free domain adaptation},
  author={Wang, Fan and Han, Zhongyi and Gong, Yongshun and Yin, Yilong},
  booktitle={Proceedings of the IEEE/CVF conference on computer vision and pattern recognition},
  pages={7151--7160},
  year={2022}
}

@article{tzeng2014deep,
  title={Deep domain confusion: Maximizing for domain invariance},
  author={Tzeng, Eric and Hoffman, Judy and Zhang, Ning and Saenko, Kate and Darrell, Trevor},
  journal={arXiv preprint arXiv:1412.3474},
  year={2014}
}

@inproceedings{lu2025dammfnd,
  title={DAMMFND: Domain-Aware Multimodal Multi-view Fake News Detection},
  author={Lu, Weihai and Tong, Yu and Ye, Zhiqiu},
  booktitle={Proceedings of the AAAI Conference on Artificial Intelligence},
  volume={39},
  number={1},
  pages={559--567},
  year={2025}
}

@article{park2024data,
  title={From data to discovery: recent trends of machine learning in metal--organic frameworks},
  author={Park, Junkil and Kim, Honghui and Kang, Yeonghun and Lim, Yunsung and Kim, Jihan},
  journal={JACS Au},
  volume={4},
  number={10},
  pages={3727--3743},
  year={2024},
  publisher={ACS Publications}
}

@inproceedings{lee2013pseudo,
  title={Pseudo-label: The simple and efficient semi-supervised learning method for deep neural networks},
  author={Lee, Dong-Hyun and others},
  booktitle={Workshop on challenges in representation learning, ICML},
  volume={3},
  number={2},
  pages={896},
  year={2013},
  organization={Atlanta}
}

@inproceedings{jaiswal2017multimedia,
  title={Multimedia semantic integrity assessment using joint embedding of images and text},
  author={Jaiswal, Ayush and Sabir, Ekraam and AbdAlmageed, Wael and Natarajan, Premkumar},
  booktitle={Proceedings of the 25th ACM international conference on Multimedia},
  pages={1465--1471},
  year={2017}
}

@article{fazio2020out,
  title={Out-of-context photos are a powerful low-tech form of misinformation},
  author={Fazio, Lisa},
  journal={The Conversation},
  volume={14},
  pages={1},
  year={2020}
}

@inproceedings{qi2024sniffer,
  title={SNIFFER: Multimodal Large Language Model for Explainable Out-of-Context Misinformation Detection},
  author={Qi, Peng and Yan, Zehong and Hsu, Wynne and Lee, Mong Li},
  booktitle={Proceedings of the IEEE/CVF Conference on Computer Vision and Pattern Recognition},
  pages={13052--13062},
  year={2024}
}

@inproceedings{huang2022text,
  title={Text-image de-contextualization detection using vision-language models},
  author={Huang, Mingzhen and Jia, Shan and Chang, Ming-Ching and Lyu, Siwei},
  booktitle={ICASSP 2022-2022 IEEE International Conference on Acoustics, Speech and Signal Processing (ICASSP)},
  pages={8967--8971},
  year={2022},
  organization={IEEE}
}

@article{aneja2021cosmos,
  title={Cosmos: Catching out-of-context misinformation with self-supervised learning},
  author={Aneja, Shivangi and Bregler, Chris and Nie{\ss}ner, Matthias},
  journal={arXiv preprint arXiv:2101.06278},
  year={2021}
}

@inproceedings{luo2021newsclippings,
  title={NewsCLIPpings: Automatic Generation of Out-of-Context Multimodal Media},
  author={Luo, Grace and Darrell, Trevor and Rohrbach, Anna},
  booktitle={Proceedings of the 2021 Conference on Empirical Methods in Natural Language Processing},
  pages={6801--6817},
  year={2021}
}

@inproceedings{mu2023self,
  title={Self-supervised distilled learning for multi-modal misinformation identification},
  author={Mu, Michael and Das Bhattacharjee, Sreyasee and Yuan, Junsong},
  booktitle={Proceedings of the IEEE/CVF Winter Conference on Applications of Computer Vision},
  pages={2819--2828},
  year={2023}
}

@article{papadopoulos2023red,
  title={RED-DOT: Multimodal Fact-checking via Relevant Evidence Detection},
  author={Papadopoulos, Stefanos-Iordanis and Koutlis, Christos and Papadopoulos, Symeon and Petrantonakis, Panagiotis C},
  journal={arXiv preprint arXiv:2311.09939},
  year={2023}
}

@inproceedings{muller2020multimodal,
  title={Multimodal analytics for real-world news using measures of cross-modal entity consistency},
  author={M{\"u}ller-Budack, Eric and Theiner, Jonas and Diering, Sebastian and Idahl, Maximilian and Ewerth, Ralph},
  booktitle={Proceedings of the 2020 international conference on multimedia retrieval},
  pages={16--25},
  year={2020}
}

@inproceedings{yuan2023support,
  title={Support or Refute: Analyzing the Stance of Evidence to Detect Out-of-Context Mis-and Disinformation},
  author={Yuan, Xin and Guo, Jie and Qiu, Weidong and Huang, Zheng and Li, Shujun},
  booktitle={The 2023 Conference on Empirical Methods in Natural Language Processing},
  year={2023},
}

@inproceedings{zhang2023ecenet,
  title={ECENet: Explainable and Context-Enhanced Network for Muti-modal Fact verification},
  author={Zhang, Fanrui and Liu, Jiawei and Zhang, Qiang and Sun, Esther and Xie, Jingyi and Zha, Zheng-Jun},
  booktitle={Proceedings of the 31st ACM International Conference on Multimedia},
  pages={1231--1240},
  year={2023}
}

@article{zhang2023interpretable,
  title={Interpretable Detection of Out-of-Context Misinformation with Neural-Symbolic-Enhanced Large Multimodal Model},
  author={Zhang, Yizhou and Trinh, Loc and Cao, Defu and Cui, Zijun and Liu, Yan},
  journal={arXiv preprint arXiv:2304.07633},
  year={2023}
}

@inproceedings{papadopoulos2025similarity,
  title={Similarity over Factuality: Are we making progress on multimodal out-of-context misinformation detection?},
  author={Papadopoulos, Stefanos-Iordanis and Koutlis, Christos and Papadopoulos, Symeon and Petrantonakis, Panagiotis C},
  booktitle={2025 IEEE/CVF Winter Conference on Applications of Computer Vision (WACV)},
  pages={5041--5050},
  year={2025},
  organization={IEEE}
}

@inproceedings{wang2018eann,
  title={Eann: Event adversarial neural networks for multi-modal fake news detection},
  author={Wang, Yaqing and Ma, Fenglong and Jin, Zhiwei and Yuan, Ye and Xun, Guangxu and Jha, Kishlay and Su, Lu and Gao, Jing},
  booktitle={Proceedings of the 24th acm sigkdd international conference on knowledge discovery \& data mining},
  pages={849--857},
  year={2018}
}

@article{zhang2020multimodal,
  title={Multimodal disentangled domain adaption for social media event rumor detection},
  author={Zhang, Huaiwen and Qian, Shengsheng and Fang, Quan and Xu, Changsheng},
  journal={IEEE Transactions on Multimedia},
  volume={23},
  pages={4441--4454},
  year={2020},
  publisher={IEEE}
}

@article{gu2024learning,
  title={Learning Domain-Invariant Features for Out-of-Context News Detection},
  author={Gu, Yimeng and Zhang, Mengqi and Castro, Ignacio and Wu, Shu and Tyson, Gareth},
  journal={arXiv preprint arXiv:2406.07430},
  year={2024}
}

@article{ganin2016domain,
  title={Domain-adversarial training of neural networks},
  author={Ganin, Yaroslav and Ustinova, Evgeniya and Ajakan, Hana and Germain, Pascal and Larochelle, Hugo and Laviolette, Fran{\c{c}}ois and March, Mario and Lempitsky, Victor},
  journal={Journal of machine learning research},
  volume={17},
  number={59},
  pages={1--35},
  year={2016}
}

@article{zhang2021understanding,
  title={Understanding deep learning (still) requires rethinking generalization},
  author={Zhang, Chiyuan and Bengio, Samy and Hardt, Moritz and Recht, Benjamin and Vinyals, Oriol},
  journal={Communications of the ACM},
  volume={64},
  number={3},
  pages={107--115},
  year={2021},
  publisher={ACM New York, NY, USA}
}

@inproceedings{sun2020test,
  title={Test-time training with self-supervision for generalization under distribution shifts},
  author={Sun, Yu and Wang, Xiaolong and Liu, Zhuang and Miller, John and Efros, Alexei and Hardt, Moritz},
  booktitle={International conference on machine learning},
  pages={9229--9248},
  year={2020},
  organization={PMLR}
}

@inproceedings{wu2024test,
  title={Test-time domain adaptation by learning domain-aware batch normalization},
  author={Wu, Yanan and Chi, Zhixiang and Wang, Yang and Plataniotis, Konstantinos N and Feng, Songhe},
  booktitle={Proceedings of the AAAI Conference on Artificial Intelligence},
  volume={38},
  number={14},
  pages={15961--15969},
  year={2024}
}

@inproceedings{osowiechi2023tttflow,
  title={Tttflow: Unsupervised test-time training with normalizing flow},
  author={Osowiechi, David and Hakim, Gustavo A Vargas and Noori, Mehrdad and Cheraghalikhani, Milad and Ben Ayed, Ismail and Desrosiers, Christian},
  booktitle={Proceedings of the IEEE/CVF Winter Conference on Applications of Computer Vision},
  pages={2126--2134},
  year={2023}
}

@article{gandelsman2022test,
  title={Test-time training with masked autoencoders},
  author={Gandelsman, Yossi and Sun, Yu and Chen, Xinlei and Efros, Alexei},
  journal={Advances in Neural Information Processing Systems},
  volume={35},
  pages={29374--29385},
  year={2022}
}

@article{liu2021ttt++,
  title={Ttt++: When does self-supervised test-time training fail or thrive?},
  author={Liu, Yuejiang and Kothari, Parth and Van Delft, Bastien and Bellot-Gurlet, Baptiste and Mordan, Taylor and Alahi, Alexandre},
  journal={Advances in Neural Information Processing Systems},
  volume={34},
  pages={21808--21820},
  year={2021}
}

@inproceedings{chen2020simple,
  title={A simple framework for contrastive learning of visual representations},
  author={Chen, Ting and Kornblith, Simon and Norouzi, Mohammad and Hinton, Geoffrey},
  booktitle={International conference on machine learning},
  pages={1597--1607},
  year={2020},
  organization={PmLR}
}

@inproceedings{he2022masked,
  title={Masked autoencoders are scalable vision learners},
  author={He, Kaiming and Chen, Xinlei and Xie, Saining and Li, Yanghao and Doll{\'a}r, Piotr and Girshick, Ross},
  booktitle={Proceedings of the IEEE/CVF conference on computer vision and pattern recognition},
  pages={16000--16009},
  year={2022}
}

@inproceedings{komodakis2018unsupervised,
  title={Unsupervised representation learning by predicting image rotations},
  author={Komodakis, Nikos and Gidaris, Spyros},
  booktitle={International Conference on Learning Representations (ICLR)},
  year={2018}
}

@inproceedings{bhattacharjee2023conda,
  title={ConDA: Contrastive Domain Adaptation for AI-generated Text Detection},
  author={Bhattacharjee, Amrita and Kumarage, Tharindu and Moraffah, Raha and Liu, Huan},
  booktitle={Proceedings of the 13th International Joint Conference on Natural Language Processing and the 3rd Conference of the Asia-Pacific Chapter of the Association for Computational Linguistics (Volume 1: Long Papers)},
  pages={598--610},
  year={2023}
}

@inproceedings{hu2024bad,
  title={Bad actor, good advisor: Exploring the role of large language models in fake news detection},
  author={Hu, Beizhe and Sheng, Qiang and Cao, Juan and Shi, Yuhui and Li, Yang and Wang, Danding and Qi, Peng},
  booktitle={Proceedings of the AAAI conference on artificial intelligence},
  volume={38},
  number={20},
  pages={22105--22113},
  year={2024}
}

@article{van2008visualizing,
  title={Visualizing Data using t-SNE},
  author={van der Maaten, Laurens and Hinton, Geoffrey},
  journal={Journal of Machine Learning Research},
  volume={9},
  pages={2579--2605},
  year={2008}
}

@article{dimitriadis2018t,
  title={t-SNE visualization of large-scale neural recordings},
  author={Dimitriadis, George and Neto, Joana P and Kampff, Adam R},
  journal={Neural computation},
  volume={30},
  number={7},
  pages={1750--1774},
  year={2018},
  publisher={MIT Press One Rogers Street, Cambridge, MA 02142-1209, USA journals-info~…}
}

@inproceedings{li2021multi,
  title={Multi-source domain adaptation with weak supervision for early fake news detection},
  author={Li, Yichuan and Lee, Kyumin and Kordzadeh, Nima and Faber, Brenton and Fiddes, Cameron and Chen, Elaine and Shu, Kai},
  booktitle={2021 IEEE International Conference on Big Data (Big Data)},
  pages={668--676},
  year={2021},
  organization={IEEE}
}

@inproceedings{yue2022contrastive,
  title={Contrastive domain adaptation for early misinformation detection: A case study on covid-19},
  author={Yue, Zhenrui and Zeng, Huimin and Kou, Ziyi and Shang, Lanyu and Wang, Dong},
  booktitle={Proceedings of the 31st ACM international conference on information \& knowledge management},
  pages={2423--2433},
  year={2022}
}

@inproceedings{mosallanezhad2022domain,
  title={Domain adaptive fake news detection via reinforcement learning},
  author={Mosallanezhad, Ahmadreza and Karami, Mansooreh and Shu, Kai and Mancenido, Michelle V and Liu, Huan},
  booktitle={Proceedings of the ACM web conference 2022},
  pages={3632--3640},
  year={2022}
}

@inproceedings{li2023blip,
  title={Blip-2: Bootstrapping language-image pre-training with frozen image encoders and large language models},
  author={Li, Junnan and Li, Dongxu and Savarese, Silvio and Hoi, Steven},
  booktitle={International conference on machine learning},
  pages={19730--19742},
  year={2023},
  organization={PMLR}
}

@inproceedings{liu2021visual,
  title={Visual News: Benchmark and Challenges in News Image Captioning},
  author={Liu, Fuxiao and Wang, Yinghan and Wang, Tianlu and Ordonez, Vicente},
  booktitle={Proceedings of the 2021 Conference on Empirical Methods in Natural Language Processing},
  pages={6761--6771},
  year={2021}
}

@article{burgess2018understanding,
  title={Understanding disentangling in $beta$-VAE},
  author={Burgess, Christopher P and Higgins, Irina and Pal, Arka and Matthey, Loic and Watters, Nick and Desjardins, Guillaume and Lerchner, Alexander},
  journal={arXiv preprint arXiv:1804.03599},
  year={2018}
}

@inproceedings{hershey2007approximating,
  title={Approximating the Kullback Leibler divergence between Gaussian mixture models},
  author={Hershey, John R and Olsen, Peder A},
  booktitle={2007 IEEE International Conference on Acoustics, Speech and Signal Processing-ICASSP'07},
  volume={4},
  pages={IV--317},
  year={2007},
  organization={IEEE}
}

@article{farahani2021brief,
  title={A brief review of domain adaptation},
  author={Farahani, Abolfazl and Voghoei, Sahar and Rasheed, Khaled and Arabnia, Hamid R},
  journal={Advances in data science and information engineering: proceedings from ICDATA 2020 and IKE 2020},
  pages={877--894},
  year={2021},
  publisher={Springer}
}

@inproceedings{you2019universal,
  title={Universal domain adaptation},
  author={You, Kaichao and Long, Mingsheng and Cao, Zhangjie and Wang, Jianmin and Jordan, Michael I},
  booktitle={Proceedings of the IEEE/CVF conference on computer vision and pattern recognition},
  pages={2720--2729},
  year={2019}
}

@article{kingma2013auto,
  title={Auto-encoding variational bayes},
  author={Kingma, Diederik P and Welling, Max},
  journal={arXiv preprint arXiv:1312.6114},
  year={2013}
}

% \clearpage

\appendix

\section{Appendix A. Baseline Models}
Considering that the problem of domain adaptation in previous Out-of-Context misinformation detection models have not explored, we are inspired by Gu et al.~\cite{gu2024learning} and select several domain adaptation models from a closely related task—fake news detection as our baseline models.

\textbf{EANN}: Event Adversarial Neural Network~\citep{wang2018eann} is a multimodal fake news detection framework that employs adversarial training.
It introduces an event discriminator to learn event-invariant features, thereby enhancing the model's generalization to unseen events.

\textbf{MDA-WS}: Multi-source Domain Adaptation with Weak Supervision~\citep{li2021multi} leverages weak supervision signals (e.g., user engagement metadata) to address the scarcity of labeled data in early fake news detection.
By integrating data from multiple source domains, it improves the model's generalization to new target domains.

\textbf{CANMD}: Contrastive Adaptation Network for Early Misinformation Detection~\citep{yue2022contrastive} aligns feature representations between source and target domains through contrastive learning, while also employing domain adversarial training to reduce intra-class variance and enlarge inter-class differences, ultimately boosting generalization to unseen domains.

\textbf{REAL-FND}: Reinforced Adaptive Learning for Fake News Detection~\citep{mosallanezhad2022domain} enhances cross-domain detection performance by dynamically optimizing feature alignment strategies.
Its core innovation is modeling domain adaptation as a Markov Decision Process and using reinforcement learning to adaptively select the best alignment strategy, effectively mitigating negative transfer in traditional adversarial training.

\textbf{CADA}: Class-based Adversarial Domain Adaptation~\citep{mosallanezhad2022domain} introduces class-aware adversarial training to preserve both domain-invariant and class-discriminative features, addressing the common issue of class confusion in standard adversarial learning.

\textbf{ConDA-TTT}: Contrastive Domain Adaptation with Test-Time Training~\citep{gu2024learning} aligns cross-modal representations between source and target domains using contrastive learning, and further incorporates a test-time training mechanism to dynamically adjust model parameters based on the target domain distribution.
This method shows strong robustness to domain shifts and unseen news events.

% \section{Appendix B. Evaluation Metrics}
% For evaluation metrics, we report accuracy and macro $F1$ score.
% Alongside the main metrics, we further include $F1_{real}$ and $F1_{fake}$~\citep{hu2024bad} to better understand class-wise performance.

% $F1_{real}$ refers to the $F1$ score of the model on the real information class.
% $F1_{fake}$ refers to the $F1$ score on the "fake information" (out-of-context misinformation) class.
% These two metrics evaluate the model’s performance on each class, helping to identify whether the model is biased toward a particular class.
% $F1_{real}$ measures the classification performance on real (non-OOC) samples.
% A low $F1_{real}$ score suggests that the model tends to be overly skeptical, with a bias toward the negative class.
% $F1_{fake}$ assesses the model’s capability to detect OOC samples.
% A low $F1_{fake}$ score indicates that the model is easily misled or insufficiently sensitive to contextual anomalies.

\section{Appendix B. Compare to the State-Of-The-Art}
In the comparison with existing models, in addition to analyzing the overall $F1$ and Accuracy (Acc), we also examine $F1_{real}$ and $F1_{fake}$. These two metrics evaluate the model’s performance on each class, helping to identify whether the model is biased toward a particular class.

Under the U,W $\rightarrow$ B setting, most methods achieve significantly higher scores on the fake class than on the real class, indicating that these models tend to misclassify real information as fake.
In contrast, our proposed VDT model achieves the highest $F1_{real}$ score of 74.08 and the smallest gap between $F1_{real}$ and $F1_{fake}$, demonstrating its strong generalization ability and under domain shift.

Under the U,W $\rightarrow$ G and B,G$\rightarrow$W settings, most models still favor the fake class, while ConDA-TTT overfits the real class, resulting in a severe drop in performance on fake class recognition.
But Our VDT model, shows a more balanced performance, indicating better robustness.

Overall, VDT maintains the smallest deviation between $F1_{real}$ and $F1_{fake}$, showing minimal bias toward either class and demonstrating excellent class balance capability.

\section{Appendix C. Statistical Significance Analysis}

\begin{figure}[]
    \centering
    \includegraphics[scale=0.5]{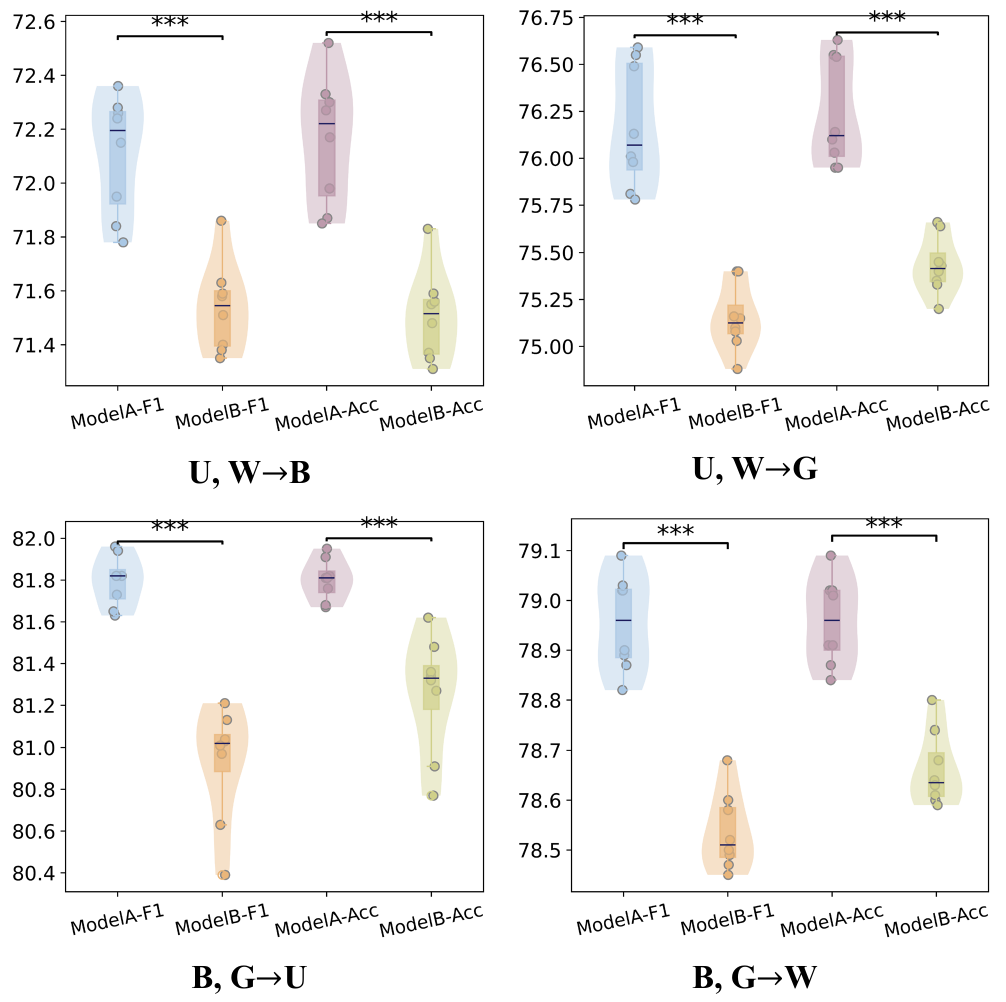}
    \caption{Statistical Comparison of performance with statistical significance annotation. "ModalA" means our proposed method VDT, "ModelB" refer the ConDA-TTT. And "*" denotes $p<0.05$, "**" denotes $p<0.01$, "***" present $p<0.001$, and "ns" for not significant.}
    \label{fig:significant}
\end{figure}

To assess the stability and significance of the performance differences, we conducted multiple runs of each model under 8 different random seeds for ConDA-TTT and our method VDT.
The performance distributions were visualized using boxplots and violin plots. 
To assess statistical significance, we also applied the Wilcoxon signed-rank test to compare paired results between models.

As shown in figure~\ref{fig:significant}, our proposed model VDT consistently demonstrates higher median values in both $F1$ and Accuracy.
Although VDT exhibits slight fluctuations in results under the setting where UW is used as the source domain, it consistently maintains a higher median across all settings.
Moreover, under the B,G$\rightarrow$ U setting, VDT shows very low variance in results, indicating more stable performance.
The paired statistical test confirms the performance improvements are statistically significant ($p < 0.001$), which indicates that the performance differences among the models are statistically significant.

\section{Appendix D. Sensitivity analysis of CVF}

To evaluate the robustness of the proposed confidence-variance filtering mechanism in the test-time training stage, we conduct a sensitivity analysis on its two key parameters $\alpha_1, \alpha_2$.
Specifically, we vary the scaling factors of confidence and the variance to observe their influence on model adaptation performance.
The experimental results are shown in Table~\ref{tab:alpha}.

\begin{table}[]
\setlength\tabcolsep{1pt}
\begin{tabular}{ccccc}
\hline
\multirow{2}{*}{$\alpha_1, \alpha_2$} & \multicolumn{1}{l}{U,W$\rightarrow$B} & \multicolumn{1}{l}{U,W$\rightarrow$G} & \multicolumn{1}{l}{B,G$\rightarrow$U} & \multicolumn{1}{l}{B,G$\rightarrow$W} \\ \cline{2-5} 
& $F1$/Acc         & $F1$/Acc         & $F1$/Acc                & $F1$/Acc                   \\  \hline
2, -1    & \textbf{72.26/72.30}  & 76.59/76.63  & \textbf{81.96/81.95}  & \textbf{79.02/79.02}   \\
1, 1     & 72.09/72.40           & 76.43/76.49           & 80.88/80.89           & 78.73/78.73       \\
1, 0     & 71.95/72.07           & \textbf{76.65/76.68}    & 81.72/81.73           & 78.86/78.90       \\
0, 1     & 71.80/72.23           & 76.60/76.65           & 80.89/80.90           & 78.66/78.67       \\
w/o gate    & 71.28/71.36      & 76.05/76.05         & 80.75/80.78              & 78.18/78.23    \\  \hline
\end{tabular}
\caption{Performance (\%) of different settings of $\alpha s$ on NewsCLIPpings. In X $\rightarrow$ Y, X denotes the source domain, Y denotes the target domain.}
\label{tab:alpha}
\end{table}

The results demonstrate that the original setting (2, –1) consistently achieves the best or near-best performance across all four domain settings.
This configuration emphasises high confidence (positive weight) while penalizing high variance (negative weight). 
The strong performance indicates that samples with confident and stable predictions indeed provide more reliable gradients during test-time training stage.
When both confidence and variance are positively weighted (1, 1), the gate mechanism may incorrectly assign high importance to uncertain samples (high variance), weakening the adaptation signal.
This explains the slight performance drop.
Both single-factor filtering (1, 0) and (0, 1) maintain reasonable performance, but neither outperforms the combined setting.
This demonstrates that confidence and variance capture complementary aspects of sample reliability, and omitting either one reduces adaptation effectiveness.
However, removing the gate leads to the most obvious degradation.

In summary, the experimental results show that combining high confidence and low variance produces the most effective sample selection strategy for test-time training, while the mechanism exhibits good robustness to parameter variations.

\section{Appendix E. MMD Distance Comparison}

To evaluate the effectiveness of the proposed DIVA module in reducing domain discrepancy, we measure the Maximum Mean Discrepancy (MMD) between the source-domain and target-domain feature distributions.
Specifically, we compute the MMD values before and after applying the DIVA module during model training.
Four domain-transfer pairs (UW$\rightarrow$B, UW$\rightarrow$G, BG$\rightarrow$U, BG$\rightarrow$W) from the NewsCLIPpings dataset are included for evaluation.
This experiment aims to verify whether the proposed DIVA module in VDT can effectively align the feature spaces of source and target domains, thereby facilitating better domain-invariant representation learning.

\begin{table}[]
\centering
\begin{tabular}{ccccc}
\hline
MMD    & UW$\rightarrow$B    & UW$\rightarrow$G    & BG$\rightarrow$U    & BG$\rightarrow$W    \\ \hline
before & 0.0607  & 0.0464  & 0.0529  & 0.0463  \\
after  & 1.72e-4 & 1.18e-4 & 1.97e-6 & 3.68e-4 \\ \hline
\end{tabular}
\caption{The MMD distance between source-domain and target-domain feature distributions. In X $\rightarrow$ Y, X denotes the source domain, Y denotes the target domain.}
\label{tab:MMD}
\end{table}

Table~\ref{tab:MMD} presents the MMD values before and after applying the DIVA module.
It can be known from the experimental results that MMD is dramatically reduced across all domain pairs.
 % And despite differences in dataset difficulty and distribution shift, all four domain pairs show similar levels of reduction, indicating that the module is domain-agnostic and not tailored to a particular source–target pair.
 The domain-invariant variational align module is designed to enforce shared semantic structure between source and target domains, the empirical reduction in MMD directly supports this theoretical claim.

 \section{Appendix F. Evaluating Dependency on Source Domain}
\begin{figure}[]
    \centering
    \includegraphics[scale=0.56]{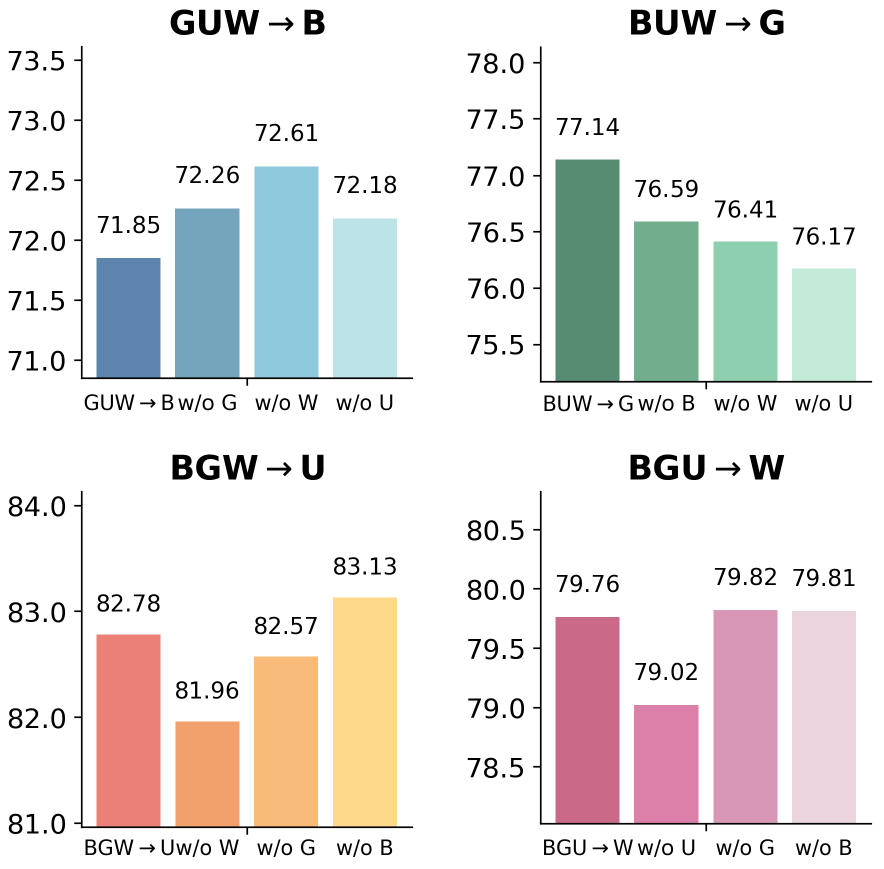}
    \caption{The $F1$ score comparison of the model after removing the source domain in sequence. In X$\rightarrow$Y, X denotes the source domains, Y denotes the target domain. And "w/o B" means removes "B" (i.e. bbc) domain data.}
    \label{fig:domain+-}
\end{figure}

To investigate whether our model relies excessively on any specific source domain, we conduct a source domain ablation study.
Under the original multi-source setting (use all domains except the target domain as the source domain), the model is trained normally.
Then, we removes one source domain while keeping all other settings identical.
For each configuration, we evaluate the final performance on all target domain setups, reporting $F1$ score in figure~\ref{fig:domain+-}.

The experimental results across the four target-domain configurations demonstrate that the model maintains stable performance even when any single source domain is excluded.
In the GUW$\rightarrow$B setting, performance slightly improves upon removing G, U, or W, suggesting the presence of mild domain interference rather than strong dependency.
For BUW$\rightarrow$G, while the full configuration achieves 77.14, removing any one domain leads only to marginal reductions (0.3–1.0), still within a non-critical range.
For BGW$\rightarrow$U, performance fluctuations remain within a narrow band (82.78 to 81.96–83.13), indicating that no individual domain is indispensable.
A similar trend is observed for BGU$\rightarrow$W, where removing B, G, or U changes the $F1$ score by less than 0.8.

Overall, the consistently small performance variations across all ablation settings indicate that the model does not overfit to or rely disproportionately on any single source domain. 
Instead, it effectively leverages information from all available sources while maintaining generalization capability when one domain is absent. 

\end{document}